%% file: main.tex
\documentclass[nohyperref]{article}

\usepackage{microtype}
\usepackage{graphicx}
\usepackage{subcaption}
\usepackage{booktabs} 

\usepackage{hyperref}
\usepackage{xurl}

\usepackage[accepted]{icml2023}

\usepackage{amsmath}
\usepackage{amssymb}
\usepackage{mathtools}
\usepackage{amsthm}

\usepackage[capitalize,noabbrev]{cleveref}

\theoremstyle{plain}

\theoremstyle{definition}

\theoremstyle{remark}

\usepackage[textsize=tiny]{todonotes}

\icmltitlerunning{Shortcut Detection with Variational Autoencoders}

\begin{document}

\twocolumn[
\icmltitle{Shortcut Detection with Variational Autoencoders}



\icmlsetsymbol{equal}{*}

\begin{icmlauthorlist}
\icmlauthor{Nicolas M. Müller}{equal,aisec}
\icmlauthor{Simon Roschmann}{equal,aisec,tum}
\icmlauthor{Shahbaz Khan}{equal,aisec,tum}
\icmlauthor{Philip Sperl}{aisec}
\icmlauthor{Konstantin B\"ottinger}{aisec}
\end{icmlauthorlist}

\icmlaffiliation{aisec}{Fraunhofer AISEC, Germany}
\icmlaffiliation{tum}{Technical University of Munich, Germany}

\icmlcorrespondingauthor{Nicolas M. Müller}{nicolas.mueller@aisec.fraunhofer.de}
\icmlcorrespondingauthor{Simon Roschmann}{simon.roschmann@tum.de}
\icmlcorrespondingauthor{Shahbaz Khan}{shahbaz.khan@tum.de}



\icmlkeywords{Shortcut Detection, Shortcuts, Spurious Correlations, Variational Autoencoder, VAE}

\vskip 0.3in
]



\printAffiliationsAndNotice{\icmlEqualContribution} 

\input{content}
\clearpage
\bibliography{references}
\bibliographystyle{icml2023}

\newpage
\appendix
\onecolumn
\input{appendix.tex}

\end{document}

%% file: content.tex
\begin{abstract}
For real-world applications of machine learning (ML), it is essential that models make predictions based on well-generalizing features rather than spurious correlations in the data.
The identification of such spurious correlations, also known as shortcuts, is a challenging problem and has so far been scarcely addressed.
In this work, we present a novel approach to detect shortcuts in image and audio datasets by leveraging variational autoencoders (VAEs).
The disentanglement of features in the latent space of VAEs allows us to discover feature-target correlations in datasets and semi-automatically evaluate them for ML shortcuts. 
We demonstrate the applicability of our method on several real-world datasets and identify shortcuts that have not been discovered before. The code is available at \href{https://github.com/Fraunhofer-AISEC/shortcut-detection-vae}{github.com/Fraunhofer-AISEC/shortcut-detection-vae}.
\end{abstract}

\section{Introduction}\label{s:intro}
Machine learning (ML) addresses a wide range of real-world problems such as quality control \cite{yang2020using}, medical diagnosis \cite{rajpurkar2022ai} and facial recognition \cite{adjabi2020past}.
However, transferring new ML technology from the lab to the real world is often difficult due to the limited capacity of ML models to generalize. Among the reasons for this limitation are shortcuts: features in data $X$ that correlate only statistically with the target $Y$, but are inconsequential for the specific ML task.
Geirhos et al. \yrcite{geirhos2020shortcut} define shortcuts as a certain group of decision rules learned by neural networks. Shortcuts perform well on training data and on independent and identically distributed (i.i.d.) test data but fail on out-of-distribution (o.o.d.) data. 

Shortcut learning has been particularly evident in the medical field. A recent MIT technology report \cite{Hundreds89:online} reveals that hundreds of tools were developed during the COVID-19 pandemic to diagnose the disease from chest X-rays, but none of them was found reliable enough for clinical use. 
The predictions of these models were often not based on the appearance of the lungs in the X-ray images. As the datasets were acquired from different sources for positive and negative cases, most of the models ended up learning the systematic differences in the data, e.g. the pose of the patient being scanned. 
Shortcut learning can also be a cause of ethical concerns towards ML. For example, due to the existence of spurious correlations in the training data, ML models have been found to reinforce gender stereotypes \cite{bolukbasi2016man, gender_bias_resume:online}.

Detecting shortcuts is a major challenge to ensure the reliability and fairness of artificial intelligence (AI).
Existing approaches \cite{zech2018confounding, singla2021causal} often rely on heatmaps and are thus limited to the identification of spatial shortcuts. We propose a novel method that is capable of identifying a variety of spurious features in images including background, color, object zoom level, and human facial characteristics.
To learn representations robust to spurious correlations, Zhang et al. \yrcite{zhang2022correctncontrast} propose a contrastive approach which ensures that the hidden representations for samples of the same class are close to each other. Kirichenko et al. \yrcite{kirichenko2022last} suggest to retrain the last layer of a model on a small dataset without spurious correlations. Our approach can serve as a preliminary step by identifying the shortcuts to be addressed.

Concurrent to our work, Yang et al. \yrcite{yang2022chroma} introduced Chroma-VAE. The authors partition the latent space of a VAE into two subspaces where one subspace is initially trained with an appended classifier to isolate shortcuts. Subsequently, the final classifier is trained on the second, shortcut-free subspace.
While this method is appealing, we believe that a human-in-the-loop approach is required to ultimately distinguish between shortcuts and valid features. In contrast to Yang et al. \yrcite{yang2022chroma} we leverage latent space traversal in VAEs and focus on the identification of shortcuts in datasets rather than the creation of a robust classifier.

\begin{figure*}
    \centering
    \includegraphics[width=0.7\textwidth]{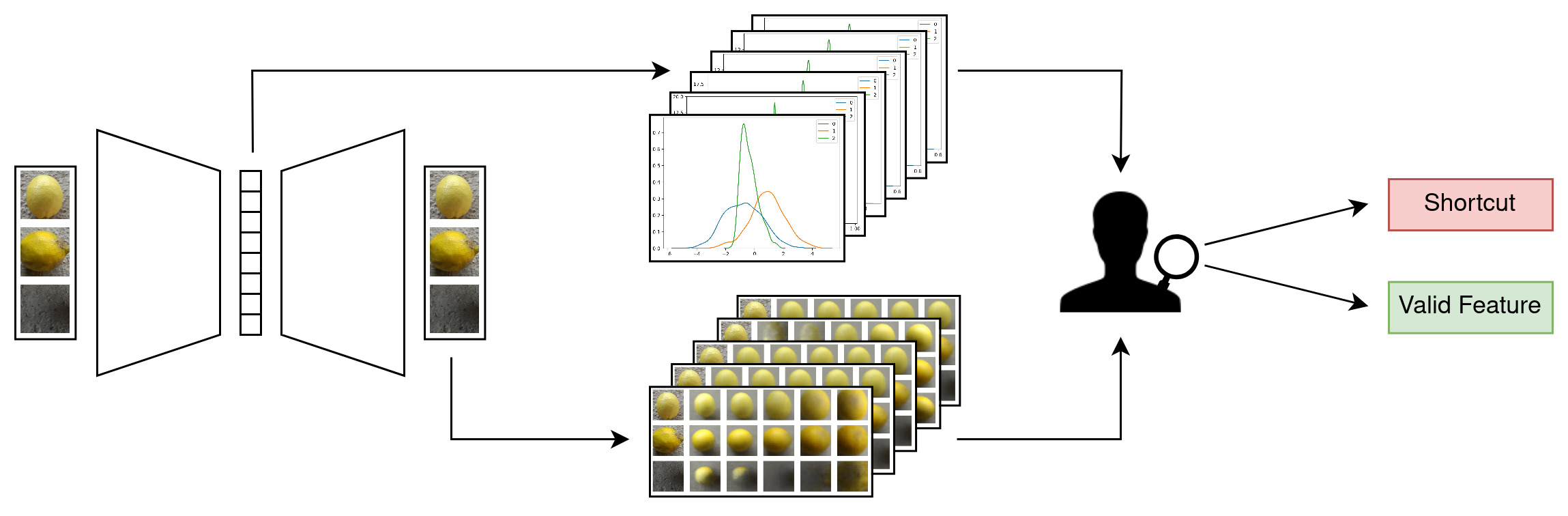}
    \caption{Shortcut detection with a VAE. We first train a Beta-VAE on a dataset containing potential shortcuts. The model discovers independent latent factors in the data and encodes them in its latent dimensions. 
    Evaluating the distribution in the latent space (top) and the weights of a VAE classifier yields a set of candidate dimensions with high predictiveness.
    The visualization of corresponding latent space traversals (bottom) enables a human judge to identify the meaning of these candidate dimensions and evaluate them for shortcuts/valid features.}
    \label{fig: methodology}
\end{figure*}

We utilize a VAE to identify correlations between features of input $X$ and target $Y$ in a dataset $D =(X, Y)$. We provide tools for visualization and statistical analysis on the latent space of the VAE which allow a human judge to reliably detect ML shortcuts as those correlations that are not meaningful to the task at hand.
We demonstrate the applicability of our approach by finding real-world shortcuts in publicly available image and audio datasets.


\section{Methodology}\label{s:approach}
Let $D = (X, Y)$ be a dataset, where $X = \{x^{(i)} | x^{(i)} \in \mathbb{R}^{h \times w \times 3}\}_{i=1}^N$ are the images, $Y = \{y^{(i)} | y^{(i)} \in \{1, ..., C\}\}_{i=1}^N$ are the corresponding targets and $C$ is the number of classes.

We train a Beta-VAE \cite{higgins2016beta} on such a dataset with potential shortcuts. The model discovers independent latent factors in a dataset and encodes them in its latent dimensions. Hence, each latent variable $z_j$ of a trained Beta-VAE is likely to represent a descriptive property of the data, e.g. brightness, color, orientation, or shape \cite{higgins2016beta}. 
To establish feature-target correlations in the data, we measure how predictive each latent variable $z_j$ is for each value of the target variable $y$ (see \Cref{ss: identification of feature-target correlations}). We perform a statistical analysis on the latent space of a VAE and assess the utility of its latent dimensions for linear classification.
For latent variables $z_j$ that show a strong correlation with the target variable $y$, we create visualizations that allow a human judge to easily decide whether the property encoded by $z_j$ is a valid feature or a shortcut (see  \Cref{ss: evaluation of feature-target correlations}).

\subsection{Identification of Feature-Target Correlations}\label{ss: identification of feature-target correlations}

Forwarding an image $x^{(i)}$ through the encoder of a trained Beta-VAE yields the parameters $\mu^{(i)}$and $\sigma^{(i)}$ of the posterior $q_{\phi}(z|x^{(i)}) = \mathcal{N} (z; \mu^{(i)}, (\sigma^{(i)})^2 I)$. The mean $\mu^{(i)}$ can be considered as the latent representation for $x^{(i)}$. We forward the entire dataset through the trained VAE, obtaining $\mu^{(i)}$ for all $i \in \{1, \ldots, N\}$.
We utilize these representations to determine the feature-target correlations in the data. We propose two different methods for this analysis.


\noindent \textbf{Statistical Analysis on the Latent Space.}
For each latent dimension $j$ and all target classes $c \in \{1, ..., C\}$, we analyze the distributions $p(z_j|y=c)$.
We are interested in finding those dimensions $j$ where two different classes result in two highly disparate distribution estimates. This indicates that feature $z_j$ is highly correlated with the target classes (a necessary, but not sufficient requirement for a shortcut).
The separation between any two distributions is quantified in terms of the Wasserstein distance \cite{vaserstein1969markov} between them.
We hypothesize that the maximum pairwise Wasserstein distance (MPWD) for a particular dimension represents its capability to separate classes.


\noindent \textbf{Classification with VAE Encoder.}
Another approach to understanding the correlation between $z_j$ and $y$ is to employ a classifier to predict $y$ given $z$. We pick the encoder of our trained Beta-VAE and append a fully connected layer. After freezing the encoder backbone,  we optimize its dense classification head. The weights in the last layer of this model denote the correlation between the latent dimensions and the target classes. 
For a dense classification head $h(z) \to y$, we define the predictiveness of a feature $z_j$ as
\begin{align}
    \text{pred}(z_j) = \sum_c |\theta_{jc}|
\end{align}
where $\theta_{jc}$ is the weight of the neuron in  $h$ which maps input $z_j$ to output $h(z_j)_c$.

\subsection{Evaluation of Feature-Target Correlations for Shortcuts}\label{ss: evaluation of feature-target correlations}
After identifying the most predictive features in the latent space, we can now evaluate them for shortcuts. To facilitate the distinction between valid and spurious correlations, we provide a human judge with 
visualizations that convey the meaning of the features.

\begin{figure}
    \centering
    \includegraphics[width=0.49\textwidth]{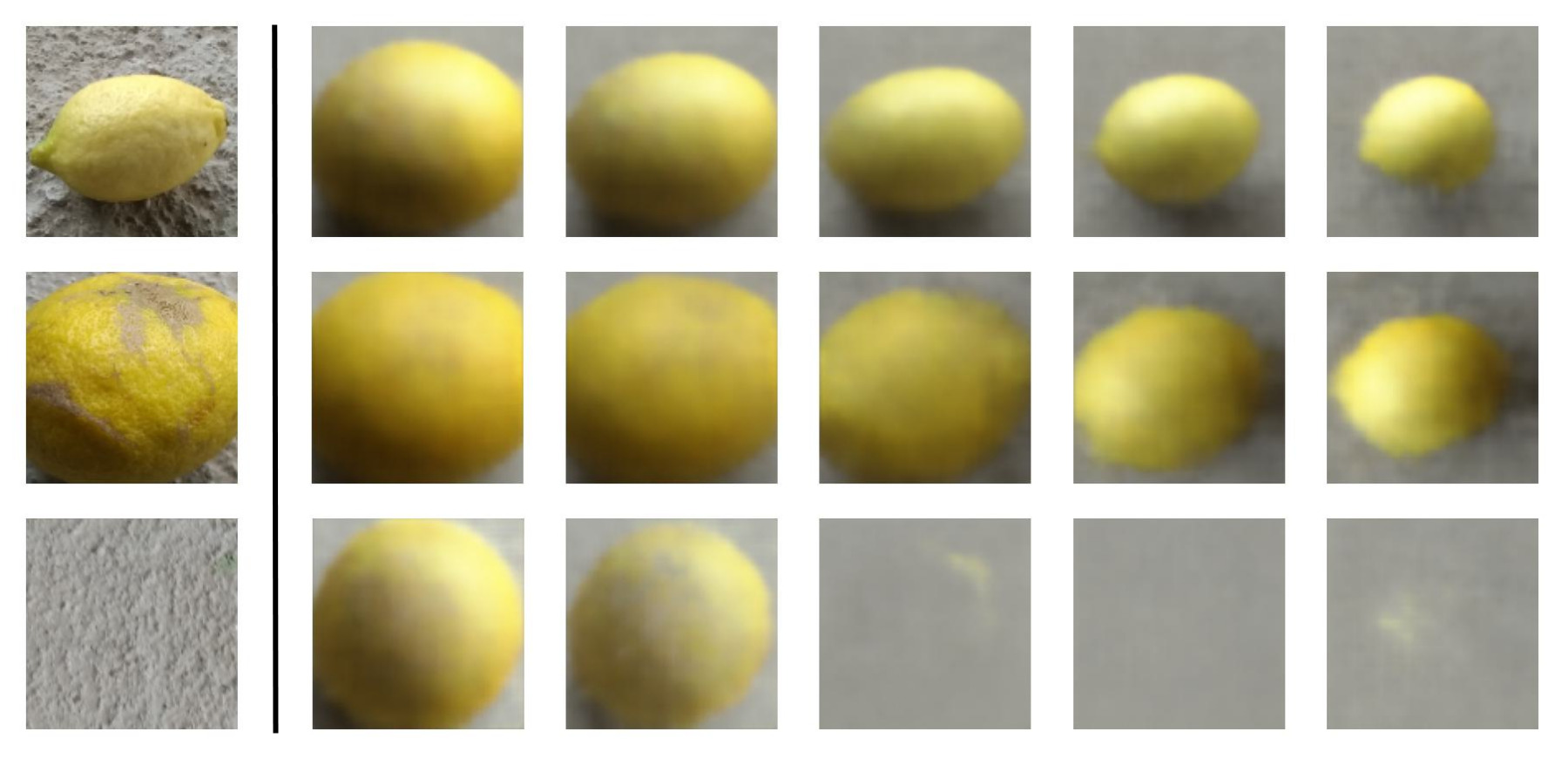}
    \caption{Latent space traversal. 
    The first column shows examples for the three classes (`good quality', `bad quality', `empty background') from the \emph{Lemon Quality} dataset.
    We vary the value of dimension $2$ in the latent space from $z_{2}^{min}=-3.75$ (second column) to $z_{2}^{max}=3.64$ (rightmost column) while keeping the values of the other dimensions fixed. Decoding the new latent representation reveals that dimension $2$ encodes the zoom level of the images. While close-up shots correlate with class `bad quality', distant shots correlate with class `good quality'.}
    \label{fig:visualization_latent_traversal}
\end{figure}

\noindent \textbf{Visualizing $z_j$ via Latent Space Traversal}. The first approach to understanding the high-level features encoded in a latent dimension $z_j$ is to visualize the effect of traversal in that dimension on the decoder output using the trained Beta-VAE.
Given an instance $x^{(i)}$, we obtain $z^{(i)} = \mu^{(i)}$ from the encoder and then visualize the decoder output $Dec(z^{(i)}+\lambda)$. The linear interpolation in the latent space is specified by $\lambda$ where $\lambda_j \in [z_j^{min}, z_j^{max}]$, $z_j^{min}=\min(\{z_j^{(i)}\}_{i=1}^N)$, $z_j^{max}=\max(\{z_j^{(i)}\}_{i=1}^N)$ and $\lambda_k = 0$ for $k \neq j$. 
This helps us to understand whether the latent variable $z_j$ encodes a useful feature or a shortcut.
\Cref{fig:visualization_latent_traversal} illustrates the latent traversal for the \emph{Lemon Quality} dataset.

\noindent \textbf{Visualizing Images for Extreme $z_j$}. To validate the meaning attributed to $z_j$, we propose to additionally compute the embeddings $z_j^{(i)}$ for all the instances $x^{(i)}$ in the dataset, and identify those instances which minimize or maximize $z_j$. This step is outlined in the Appendix.

\subsection{On the Necessity of Human Judges}\label{ss:human_intervention_necessity}
Our method requires a human judge to evaluate if the information encoded by the latent variable $z_j$ constitutes a feature or a shortcut. We argue that human interventions are inevitable for ML-based shortcut detection \cite{zech2018confounding, geirhos2020shortcut, singla2021causal}. ML models learn relations between input and output based on statistics alone. Unlike humans, they are hardly equipped with prior knowledge of the real world beyond the given dataset and the specified task. However, this prior knowledge may be necessary to evaluate whether a correlation is valid or spurious. Therefore, to assess the candidate feature-target correlations identified by our model, we need a human judge in the last step of our approach. 
This human supervision is limited to inferring the meaning of a latent dimension from the visualization of its latent traversal.

\begin{figure}
    \centering
    \includegraphics[width=0.49\textwidth]{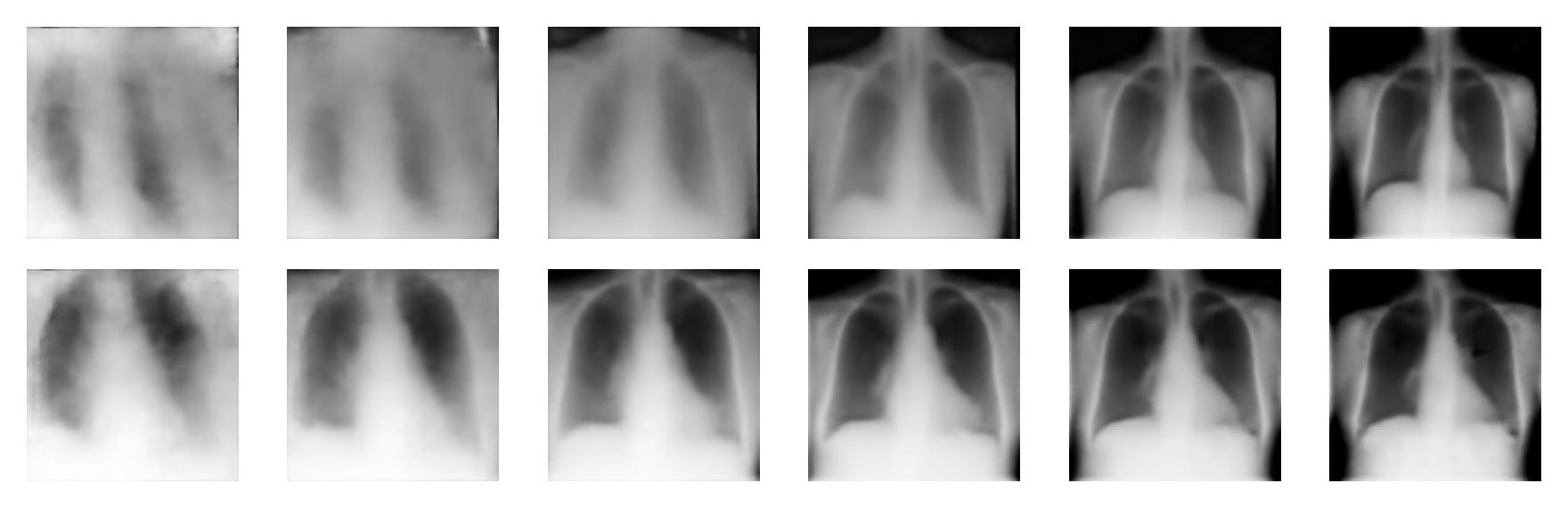}
    \caption{Evaluation of our method on the \emph{COVID-19} dataset. The position of the patient's chest in the X-ray images is revealed as a spurious attribute through latent space traversal.
    Patients with `covid' appear closer to the scanner (left) while patients with `no covid' appear further away such that the black background on the top and bottom sides is visible (right).}
    \label{fig:covid_results}
\end{figure}

\section{Evaluation}\label{s:eval}

\subsection{Experimental Setup}\label{ss:experimental_setup}
To obtain a low-dimensional latent representation, we use a Beta-VAE (see \Cref{ss:background_vae}) with $dim(z) \in \{10, 32\}$ depending on the data.
For our encoder, we employ a ResNet \cite{he2016deep} backbone pretrained on the ImageNet \cite{russakovsky2015imagenet} dataset. 
We append two separate linear layers for predicting the parameters $\mu$ and $\sigma$ of the posterior distribution.
The decoder of our model consists of $5$ hidden layers, each applying transposed convolutions with padding $1$, stride $2$, and kernel size $3$. 
The channel dimensions of the successive hidden layers in the decoder are chosen as follows: $512$, $256$, $128$, $64$, and $32$.
We apply ReLU activation and batch normalization in the hidden layers and Sigmoid activation in the output layer.
While our model can operate on varying input sizes, we use images of size $128 \times 128$ in our experiments. We do not perform any data augmentation so as to retain the original characteristics of the input data including any shortcut.

Our model is trained using the Adam \cite{kingma2014adam} optimizer with a learning rate of $0.001$ and a batch size of $32$.
Since $\beta$ determines the tradeoff between reconstruction and sampling quality, we perform hyperparameter tuning on $\beta$ for each dataset. 
To estimate the predictiveness of latent features, we append a linear layer to the frozen encoder of a trained Beta-VAE. 
The resulting classifier with cross-entropy loss is trained with the Adam optimizer and a learning rate of $0.001$ on batches of size $32$.
We ran all experiments on an Nvidia Titan X 12GB GPU.

\begin{table*}
\centering
\caption{Shortcut detection results. For every dataset, we denote the latent dimension of a trained VAE which encodes a spuriously correlated attribute. They are picked from a set of candidate dimensions shortlisted based on how predictive (see \Cref{ss: identification of feature-target correlations}) they are in relation to all dimensions. The semantic meaning encoded by these candidate dimensions is revealed with a traversal in the latent space (see \Cref{ss: evaluation of feature-target correlations}).}
\label{tab: results}
\begin{tabular}{@{}lclccc@{}}
\toprule
Dataset & Spurious Dimension & Spurious Attribute & MPWD & Predictiveness\\
\midrule
ASVspoof & 30 & Leading silence & 2/32 & 2/32\\
CelebA & 26 & Gender & 3/32 & 3/32 \\
Colored MNIST & 14 & Color & 1/32 & 1/32 \\
COVID-19 & 1 & Patient's position & 1/32 & 2/32\\
Lemon Quality & 2 & Zoom level & 1/10 & 2/10\\
Waterbirds & 1 & Background & 2/32 & 3/32 \\
\bottomrule
\end{tabular}
\end{table*}

\subsection{Datasets} \label{ss:datasets}
Our approach is evaluated on six datasets: \emph{Waterbirds} \cite{sagawa2019distributionally}, \emph{Colored MNIST} \cite{arjovsky2019invariantriskminimization}, \emph{CelebA} \cite{liu2015celeba}, \emph{Lemon Quality} \cite{lemondataset}, \emph{COVID-19} \cite{degrave2021ai}, \emph{ASVspoof} \cite{wang2020asvspoof}. Details on these datasets are provided in the Appendix.

\subsection{Results}\label{ss:results_eval}
The empirical results are summarized in \Cref{tab: results}. While the particular dimension representing a spurious feature can vary between different training runs, any VAE trained with well-chosen hyperparameters should be able to encode the feature in one of its latent dimensions.
The MPWD and the predictiveness of a spurious dimension are reported relative to all latent dimensions.
Across all experiments, we conclude that a human judge has to review only the top $k = 3$ latent dimensions with the highest MPWD and predictiveness to identify a shortcut if present in the data. 

Our approach correctly identifies the known shortcut in the \emph{COVID-19} dataset. The statistical analysis on the latent space of a trained VAE shows that its latent variable $z_{1}$ has the highest MPWD and a high predictiveness $pred(z_{1})$ (among top 2). The latent traversal illustrated in \Cref{fig:covid_results} reveals that the latent variable $z_{1}$ encodes the position of a patient.
Similarly, we detect the color shortcut in the \emph{Colored MNIST} dataset, the background shortcut in the \emph{Waterbirds} dataset and the gender shortcut in the \emph{CelebA} dataset. 
To the best of our knowledge, we are the first to identify the existence of a shortcut in the \emph{Lemon Quality} dataset. As depicted in \Cref{fig:visualization_latent_traversal}, our method reveals the correlation between lemon quality and zoom level. 
We make a step towards the generalization of our method to other domains by identifying shortcuts in spectrograms from the \emph{ASVspoof} dataset. 


To compare our results to heatmap-based approaches \cite{zech2018confounding, singla2021causal}, we generate heatmaps (see Appendix) of maximally activating features of the penultimate layer for a standard CNN classifier.
The heatmaps for the \emph{Lemon Quality} dataset focus on brown patches for class `bad quality'  and on the background for class `good quality'. This spatial shortcut is also identified by our method.
For \emph{CelebA}, we obtain heatmaps that focus on meaningful facial features, including hair, forehead and eyes. In contrast to our approach, none of these heatmaps make the gender shortcut in the dataset obvious. 

\subsection{Limitations}

Dai and Wipf \yrcite{dai2019diagnosing} have argued that latent representations with independent dimensions do not necessarily correspond to any semantically-meaningful form of disentanglement. Locatello et al. \yrcite{locatello2019challenging} have theoretically proven that the unsupervised learning of disentangled representations is impossible without inductive biases on the learning approaches and the datasets.
The human supervision in our pipeline addresses the concern of Dai and Wipf \yrcite{dai2019diagnosing}. 
Once we quantitatively discover a latent variable that shows a high correlation with a particular class label, a human judge can inspect its latent traversal.
The human-in-the-loop approach allows to determine whether the recovered visual pattern in a particular latent dimension matches a real-world concept. 

A Beta-VAE can encode a disentangled representation only if the underlying factors of a dataset are independent. The existence of shortcuts in a dataset makes it particularly difficult to fully separate the factors of variation. 
However, we demonstrate that our approach even works if two or more real-world concepts are entangled in one latent dimension. \Cref{fig:covid_results} illustrates how the change of a spurious attribute (e.g. position) can coincide with the change of the main distinguishing factor (e.g. lung appearance).
Finally, we note that VAEs are sensitive to the choice of hyperparameters, including the number of latent dimensions and the weights in the loss function. 

\section{Conclusion}\label{s:conclusion}
In this paper, we introduce a novel approach to detect spurious correlations in machine learning datasets. Our method utilizes a VAE to discover meaningful features in a given dataset and enables a human judge to identify shortcuts effortlessly.
We evaluate our approach on image and audio datasets and successfully reveal the inherent shortcuts.
We hope that our work inspires researchers to explore the potential of VAEs in the field of shortcut learning. It would be interesting to see if our method can be extended to other domains.

%% file: appendix.tex
\section{Related Work}\label{s:related_work}

\noindent\textbf{Machine Learning Shortcuts.} Geirhos et al. \yrcite{geirhos2020shortcut} locate the origin of shortcuts in the data and the learning process of ML models. The inherent contextual bias in datasets provides opportunities for shortcuts. Natural image datasets contain spurious correlations between the target variable and the background \cite{xiao2020noise}, the object poses \cite{alcorn2019strike} or other co-occurring distracting features \cite{kolesnikov2016improving, shetty2019not}.
In discriminative learning, a model uses a combination of features to make a prediction. Following the ``principle of least effort'', models tend to rely only on the most obvious features, which often correspond to shortcuts \cite{geirhos2020shortcut}.
For instance, convolutional neural networks (CNNs) trained on ImageNet were found to be biased towards the texture of the objects instead of their shapes \cite{geirhos2018imagenet}. In another example, it was discovered that CNNs solely used the location of a single pixel to distinguish between object categories \cite{malhotra2018difference}.

\noindent\textbf{Identification of Shortcuts.} 
The identification of shortcuts in supervised machine learning is still in its infancy. 
Zech et al. \yrcite{zech2018confounding} use activation heatmaps to reveal the spurious features learned by CNNs trained on X-ray images.
As outlined by Viviano et al. \yrcite{viviano2019saliency}, saliency maps can only explain spatial shortcuts, e.g. source tags on images \cite{lapuschkin2019unmasking}, but fail to identify more complicated ones, e.g. people's gender \cite{sagawa2019distributionally}.
Singla and Feizi \yrcite{singla2021causal} select the highest activations of neurons in the penultimate layer of a CNN classifier and back-project them onto the input images. 
The resulting heatmaps highlight the features in the image that maximize neural activations. 
Under human supervision, the highlighted regions, for a subset of images, are labelled as `core' (part of the object definition) or `spurious' (only co-occurring with the object). Using this labelled dataset, they train a classifier to automatically identify the core and spurious visual features for a larger dataset. 
To diagnose shortcut learning, Geirhos et al. \yrcite{geirhos2020shortcut} suggest performing o.o.d. generalization tests. Evaluating the model on o.o.d. real-world data in addition to the i.i.d. test set reveals whether a model is actually generalizing on the intended features or simply learning shortcuts from the training data.

\noindent\textbf{Robustness against Shortcuts.} 
To learn representations robust to spurious correlations, Zhang et al. \yrcite{zhang2022correctncontrast} propose a two-stage contrastive approach. The method first identifies training samples from the same class with different model predictions. Contrastive learning then ensures that the hidden representations for samples of the same class with initially different predictions become close to each other.
Kirichenko et al. \yrcite{kirichenko2022last} suggest to retrain the last layer of a classifier on a small dataset without spurious correlations. While the reweighting of the last layer reduces the model's reliance on background and texture information, the requirement of a shortcut-free subset remains a limitation of this approach. Our method could serve as a preliminary step by identifying the shortcuts to be addressed.

\section{Methodology}
\subsection{Variational Autoencoder}\label{ss:background_vae}
Our approach for shortcut detection is based on variational autoencoders \cite{kingma2013auto} (VAEs), which are probabilistic generative models that learn the underlying data distribution in an unsupervised manner.
A VAE attempts to model the marginal likelihood of an observed variable $x$:
\begin{equation}
p_{\theta}(x) = \int p_{\theta}(z) p_{\theta}(x|z) \,dz
\end{equation} \label{eq:integral}The unobserved variable $z$ lies in a latent space of dimensionality $d = \dim(z)$, $d \ll \dim(x) $. Each instance $x^{(i)}$ has a corresponding latent representation $z^{(i)}$.

The model assumes the prior over the latent variables to be a multivariate normal distribution $p_{\theta}(z) = \mathcal{N} (z; 0, I)$ resulting in independent latent factors $\{z_j\}_{j=1}^d$.
The likelihood $p_{\theta}(x|z)$ is modelled as a multivariate Gaussian whose parameters are conditioned on $z \sim p_{\theta}(z)$ and computed using the VAE decoder. 
As outlined by Kingma and Welling \yrcite{kingma2013auto}, the true posterior $p_{\theta}(z|x)$ is intractable and hence approximated with a variational distribution $q_{\phi}(z|x)$. This variational posterior is chosen to be a multivariate Gaussian

\begin{equation}\label{eq:posterior}
    q_{\phi}(z|x^{(i)}) = \mathcal{N} (z; \mu^{(i)}, (\sigma^{(i)})^2 I)
\end{equation}

whose mean $\mu^{(i)} \in \mathbb{R}^d$ and standard deviation $\sigma^{(i)} \in \mathbb{R}^d$ are obtained by forwarding $x^{(i)}$ through the VAE encoder.
The objective function is composed of two terms. A Kullback-Leibler (KL) divergence term ensures that the variational posterior distribution remains close to the assumed prior distribution. 
\begin{equation}
\begin{aligned}
    D_{KL}(q_{\phi}(z|x^{(i)})||p_{\theta}(z))
    = -\frac{1}{2}\sum_{j=1}^{d}\left( 1 + \text{log}((\sigma_j^{(i)})^2) - (\sigma_j^{(i)})^2 - (\mu_j^{(i)})^2 \right) \label{eq:kld}
\end{aligned}
\end{equation}
A log-likelihood loss, on the other hand, helps to accurately reconstruct an input image $x^{(i)}$ from a sampled latent variable $z^{(i)} = \mu^{(i)} + \sigma^{(i)} \odot \epsilon $ where $\epsilon \sim \mathcal{N}(0, I)$.
Thus the encoder and decoder are trained to maximize the following objective function:
\begin{equation}\label{eq:vae_loss}
\begin{aligned}
    \mathcal{L}(\theta, \phi; x^{(i)})
    = -D_{KL}(q_{\phi}(z|x^{(i)}) || p_{\theta}(z)) + log~p_{\phi} (x^{(i)} | z^{(i)})
\end{aligned}
\end{equation}
For modelling images, both the encoder and decoder of a VAE consist of CNNs.

Higgins et al. \yrcite{higgins2016beta} introduce the Beta-VAE to better learn independent latent factors. The authors propose to augment the vanilla VAE loss in \Cref{eq:vae_loss} by weighing the KL term with a hyperparameter $\beta$. Choosing $\beta > $  1 enables the model to learn a more efficient latent representation of the data with better disentangled dimensions. 

\subsection{Hyperparameter Tuning of Beta-VAE}
Following Higgins et al. \yrcite{higgins2016beta}, we tune the hyperparameters $dim(z)$ and $\beta$ of the Beta-VAE for every dataset. To achieve maximum disentanglement, the number of latent dimensions should match the number of factors of variation in a dataset. This number is usually not known a priori. Choosing a latent space of too high dimensionality leads to a lot of uninformative dimensions with low variance. A latent space of too few latent dimensions, on the other hand, leads to entangled representations of features in the latent space. Therefore, it is necessary to identify the optimal number of latent dimensions to achieve maximally disentangled factors, ideally one in each dimension. 

To obtain the best combination of $\beta$ and $dim(z)$ for a given dataset, we first train a Beta-VAE with $\beta=1$ and $dim(z)=32$ on all datasets. We then compare the variance of the distribution in each dimension to that of a Gaussian prior. In the presence of dimensions with relatively low variance, we reduce the number of latent dimensions.
Once we find a Beta-VAE with consistently informative dimensions, i.e. with the variance comparable to the prior, we fix $dim(z)$ and focus on fine-tuning $\beta$.
As outlined by Higgings et al. \yrcite{higgins2016beta}, for relatively low values of $\beta$, the VAE learns an entangled latent representation since the capacity in the latent space is too high. On the other hand, for relatively high values of $\beta$, the capacity in the latent space becomes too low. The VAE performs a low-rank projection of the true data generative factors and again learns an entangled latent representation. This renders some of the latent dimensions uninformative. 

We find the highest possible $\beta$ for which all dimensions of the VAE remain informative with a variance close to the prior. In line with the findings of Higgings et al. \yrcite{higgins2016beta}, $\beta > 1$ is required for all datasets to achieve good disentanglement (see~\cref{tab: hyperparameter tuning}).

\begin{table}[h!]
\centering
\caption{Beta-VAE hyperparameters}
\label{tab: hyperparameter tuning}
\begin{tabular}{@{}lcc@{}}
\toprule
Dataset & $dim(z)$ & $\beta$\\
\midrule
ASVspoof & 32 & 1.25\\
CelebA & 32 & 10.0\\
Colored MNIST & 32 & 2.5\\
COVID-19 & 32 & 1.5\\
Lemon Quality & 10 & 3.0\\
Waterbirds & 32 & 1.75\\
\bottomrule
\end{tabular}
\end{table}

\subsection{Visualizing Images for Extreme $z_j$}
\begin{figure}
\begin{subfigure}{0.49\textwidth}
\includegraphics[width=\textwidth]{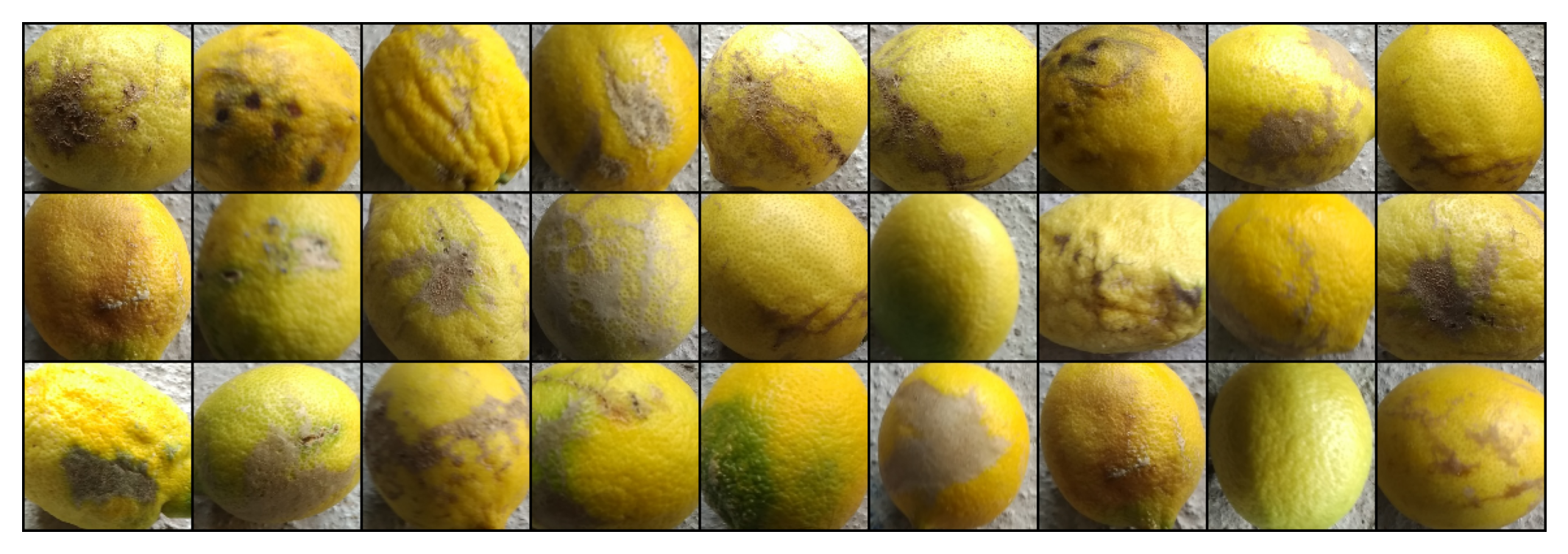}
\caption{Inputs from the \emph{Lemon Quality} dataset which minimize $z_2$.}
\label{fig:minimizer}
\end{subfigure}
\hfill
\begin{subfigure}{0.49\textwidth}
\includegraphics[width=\textwidth]{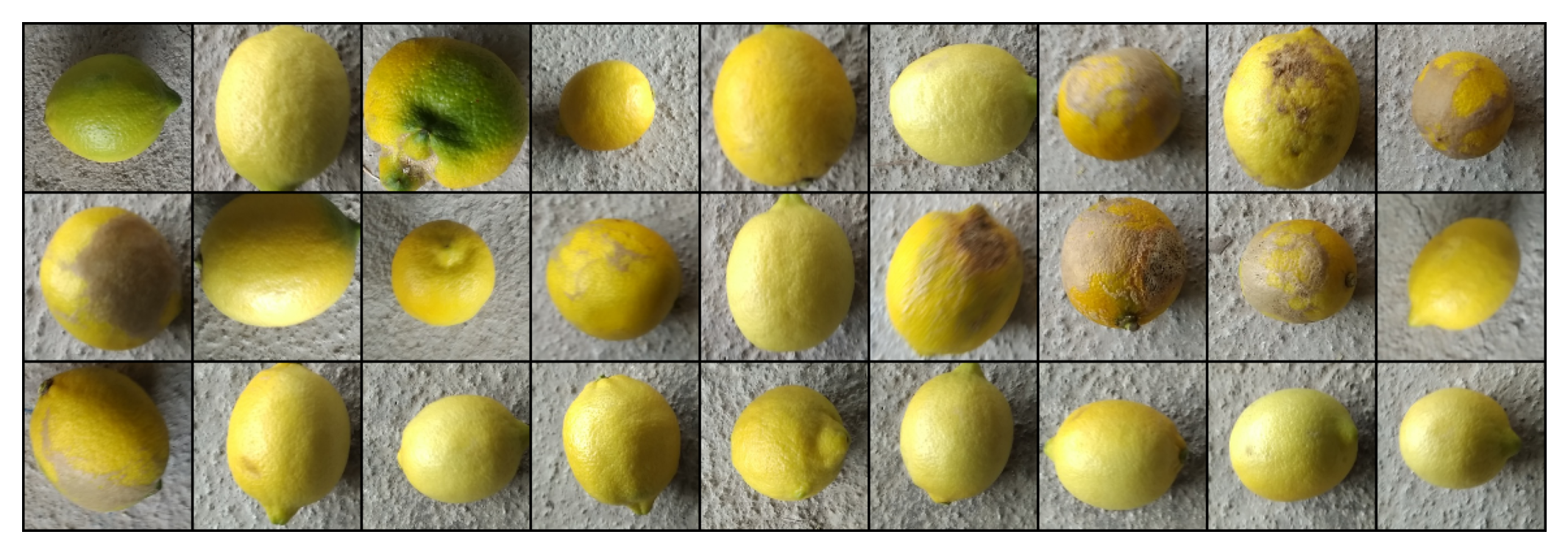}
\caption{Inputs from the \emph{Lemon Quality} dataset which maximize $z_2$.}
\label{fig:maximizer}
\end{subfigure}
\caption{Real-world images from the training dataset that correspond to the minimum and maximum encoded values in latent dimension $2$. While bad-quality lemons are mostly captured as close-up shots, good-quality lemons are photographed from a distance.}
\label{fig:visualization_existing_data}
\end{figure}

To validate the meaning attributed to $z_j$, we propose to additionally compute the embeddings $z_j^{(i)}$ for all the instances $x^{(i)}$ in the dataset, and identify those instances which minimize or maximize $z_j$.
Particularly, we perform $\arg\operatorname{sort}_{i} (\{z_j^{(i)}\}_{i=1}^N)$ for every dimension $j$ in the latent space and display the input images $x^{(i)}$ corresponding to the first $l$ and the last $l$ indices of the sorted values.
\Cref{fig:visualization_existing_data} depicts $l = 27$ images of the \emph{Lemon Quality} dataset with minimum and maximum values in latent dimension $2$. 

\section{Evaluation}
\subsection{Datasets}

We evaluate our approach on datasets with artificially introduced shortcuts to demonstrate its ability to identify spurious correlations. 
Our method can also be applied to real-world datasets to reveal previously unknown shortcuts. We perform a train-val-test split in the ratio 80:10:10 on each dataset unless explicitly specified.

\noindent\textbf{Waterbirds.} 
Sagawa et al. \yrcite{sagawa2019distributionally} extract birds from the Caltech-UCSD Birds-200-2011 dataset \cite{wah2011caltech} and combine them with background images from the Places dataset \cite{zhou2017places}. 
As a result, 95\% of the water birds appear on a water background while 95\% of the land birds appear on a land background. 
Since this spurious correlation is only introduced in the train set of the \emph{Waterbirds} dataset consisting of 4,795 samples, we use the same to create the train-val-test splits for our experiments.

\noindent\textbf{Colored MNIST.} 
Arjovsky et al. \yrcite{arjovsky2019invariantriskminimization} inject color as a spurious attribute into the MNIST \cite{lecun2010mnist} dataset. 
Following this idea, Zhang et al. \yrcite{zhang2022correctncontrast} create a colored MNIST dataset consisting of five subsets with five associated colors.
A fraction $p_{corr}$ of the training samples are assigned colors based on the group they belong to. The remaining samples are assigned a random color.
We follow the color assignment in \cite{zhang2022correctncontrast} and choose $p_{corr} = 0.995$ for coloring the 70,000 MNIST samples.

\noindent\textbf{CelebA.} 
The CelebFaces Attributes Dataset \cite{liu2015celeba} contains 202,599 images of celebrities, each annotated with 40 facial attributes.
Sagawa et al. \yrcite{sagawa2019distributionally} train a classifier to identify the hair color of the celebrities and discover that the target classes (blond, dark) are spuriously correlated with the gender (male, female). We stick to the setup specified in \cite{sagawa2019distributionally} with the official train-val-test splits of the \emph{CelebA} dataset.

\noindent\textbf{Lemon Quality.} 
To demonstrate the detection of shortcuts in quality control, we apply our method on the \emph{Lemon Quality} dataset \cite{lemondataset}. The dataset consists of 2,533 images labelled with one of three classes, namely `good quality', `bad quality', and `empty background'.

\noindent\textbf{COVID-19.} 
Following \cite{brunese2020explainable, ghoshal2020estimating, hemdan2020covidx, ozturk2020automated, degrave2021ai}, we obtain a dataset with 112,528 samples for COVID-19 detection by combining COVID-19-positive radiographs from the GitHub-COVID repository \cite{cohen2020covid} and COVID-19-negative radiographs from the ChestX-ray14 repository \cite{wang2017chestx}.

\noindent\textbf{ASVspoof.} The ASVspoof 2019 Challenge Dataset \cite{wang2020asvspoof} is used to train and benchmark systems for the detection of spoofed audio and audio deepfakes. Müller at al. \yrcite{muller2021speech} observe that the length of the silence at the beginning of an audio sample differs significantly between benign and malicious data. Previous deepfake detection models have exploited this shortcut.
We chose a subset of the training data (benign audio and attack $A01$), which results in a binary classification dataset comprising 6,380 samples. 
We transform the audio samples to CQT spectrograms~\cite{schorkhuber2010constant}, and obtain a frequency-domain representation with $257$ logarithmically spaced frequency bins, capturing up to $8$ kHz (Nyquist frequency given the input is $16$ kHz).

\subsection{Results}

We provide illustrations of the latent space traversal for the \emph{Colored MNIST}, \emph{CelebA} and \emph{ASVspoof} dataset.


\begin{figure}[h!]
    \centering
    \includegraphics[width=0.7\textwidth]{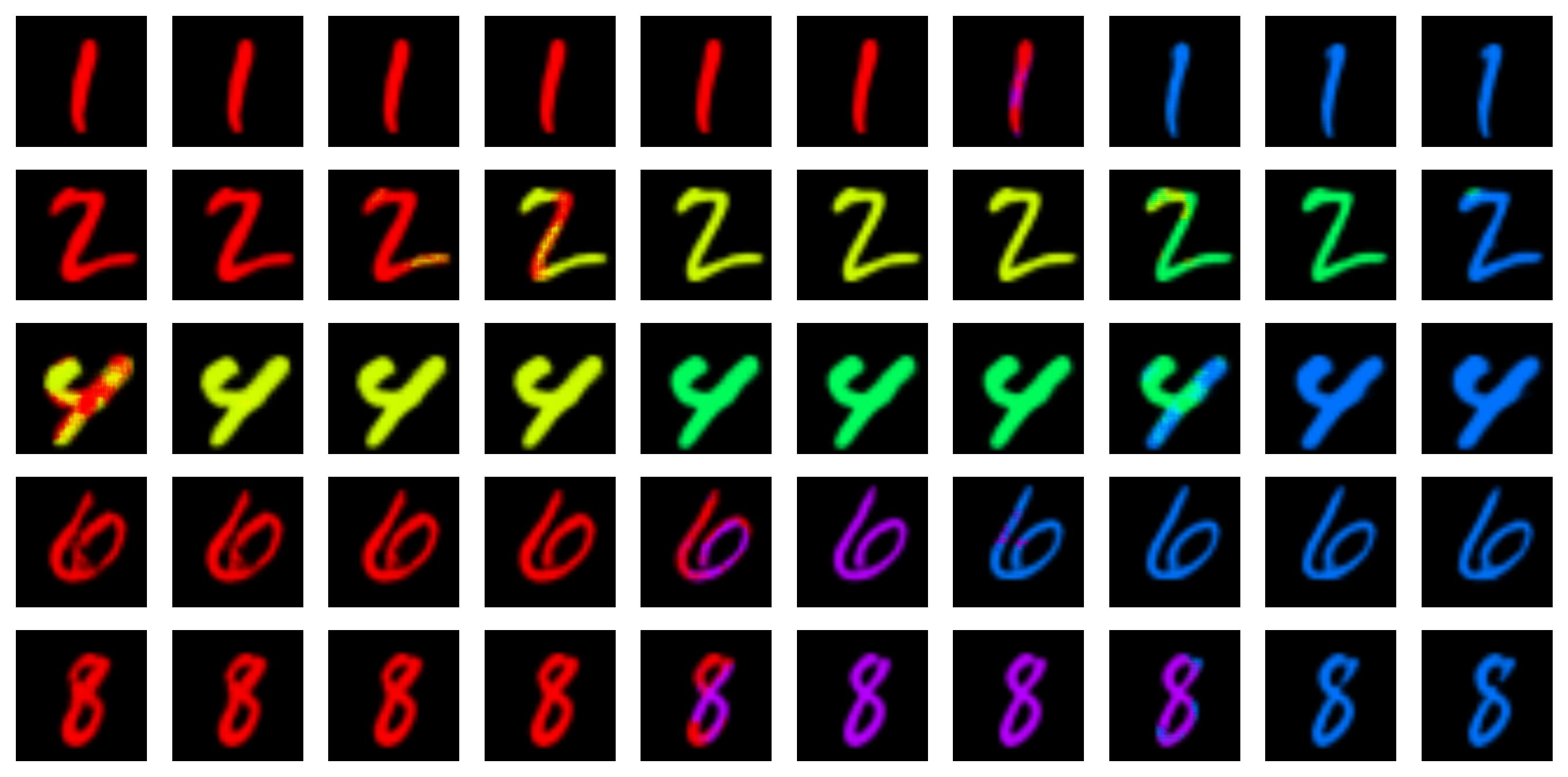}
    \caption{Evaluation of our method on the \emph{Colored MNIST} dataset.  The illustrated dimension represents the attribute `color'. The minimum values encode the colors red (left) while the maximum values encode the color blue (right).}
    \label{fig:cmnist_results}
\end{figure}

\begin{figure}[h!]
    \centering
    \includegraphics[width=0.7\textwidth]{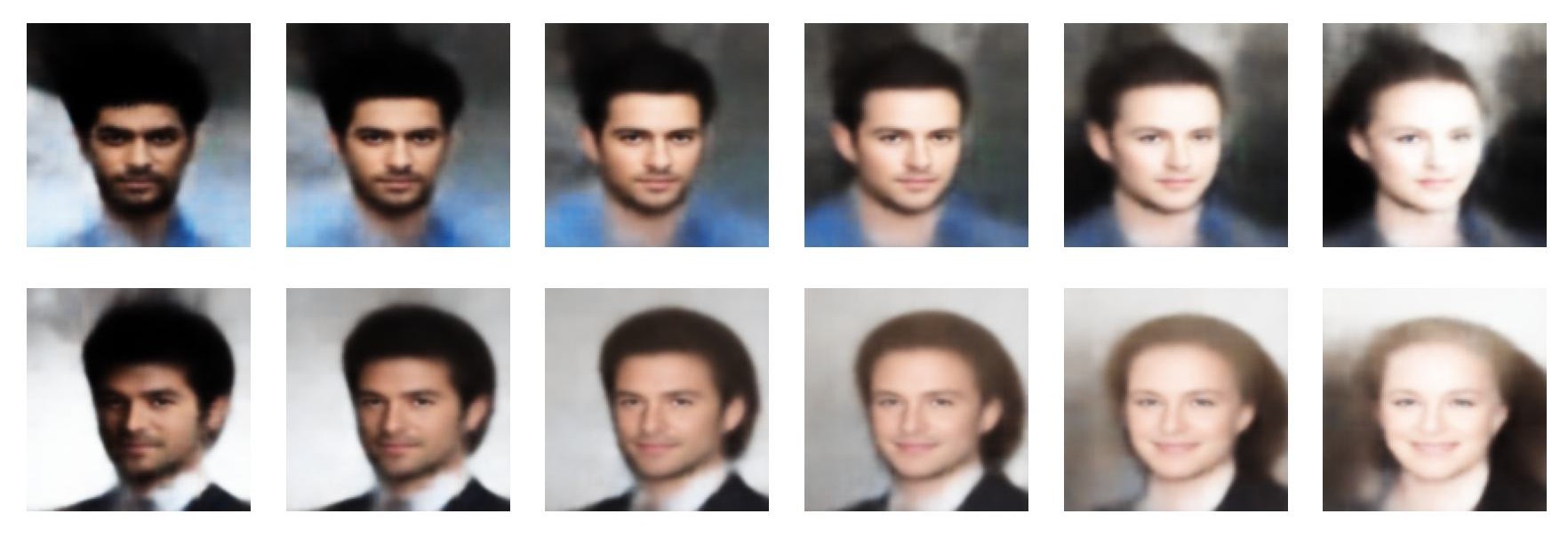}
    \caption{Evaluation of our method on the \emph{CelebA} dataset. The latent traversal reveals the meaning of dimension $26$ as `gender'. The high correlation (as measured in terms of MPWD and predictiveness) of this attribute with the target variable `hair color' indicates the existence of a spurious correlation.}
    \label{fig:celeba_gender_hair_results}
\end{figure}

\begin{figure}[h!]
    \centering
    \includegraphics[width=0.7\textwidth]{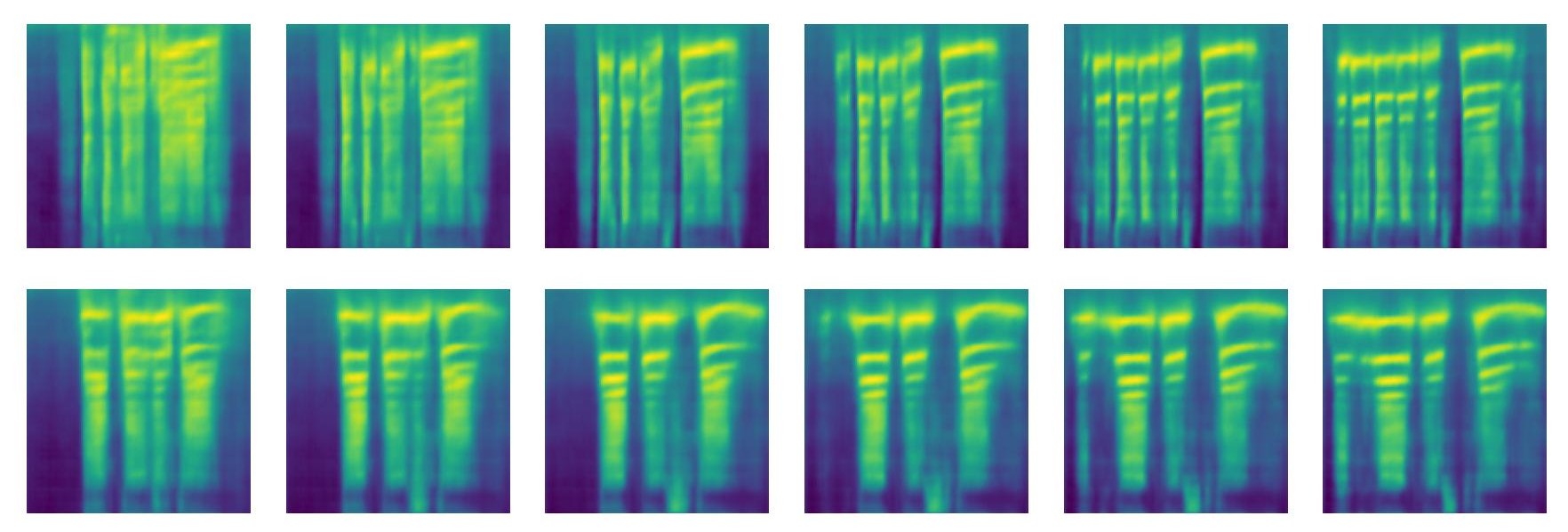}
    \caption{Evaluation of our method on the \emph{ASVspoof} dataset.
    Latent traversal reaffirms the known spurious correlation between the leading silence in the audio and the target class. Leading silence (left) in the spectrogram is an indicator of benign audio samples while no leading silence (right) is common in spoofs.}
    \label{fig:asvspoof_results}
\end{figure}


\subsection{Comparison}
To further evaluate our approach, we compare it to one of the most established Explainable AI techniques, namely heatmaps \cite{zech2018confounding, singla2021causal}.
We train a CNN $C(x) = L(F(x))$, which consists of a convolutional feature extractor $F$ and a linear model $L$, where $L$ computes a linear combination of the features $F(x)$. We design $F$ to include five convolutional layers with 32, 64, 128, 256, and 512 filters, using a kernel of size $3 \times 3$. Each of these layers is followed by ReLU activation and max pooling.

For each image $x$, we obtain features $F(x)$ in the penultimate layer. We compute the average contribution of each feature $F(x)_i$ towards the activation $L(F(x)_i)_c$ of a particular class $c$ by multiplying it with the associated weight. The average contribution of each feature to the prediction of a certain class is computed over all images of that class. We pick the top $5$ most predictive features and then find the top $5$ images that yield the highest activation. We compute the heatmaps by projecting these top features onto the original images.

\begin{figure}[h!]
    \centering
    \begin{subfigure}{0.45\textwidth}
        \includegraphics[width=\textwidth]{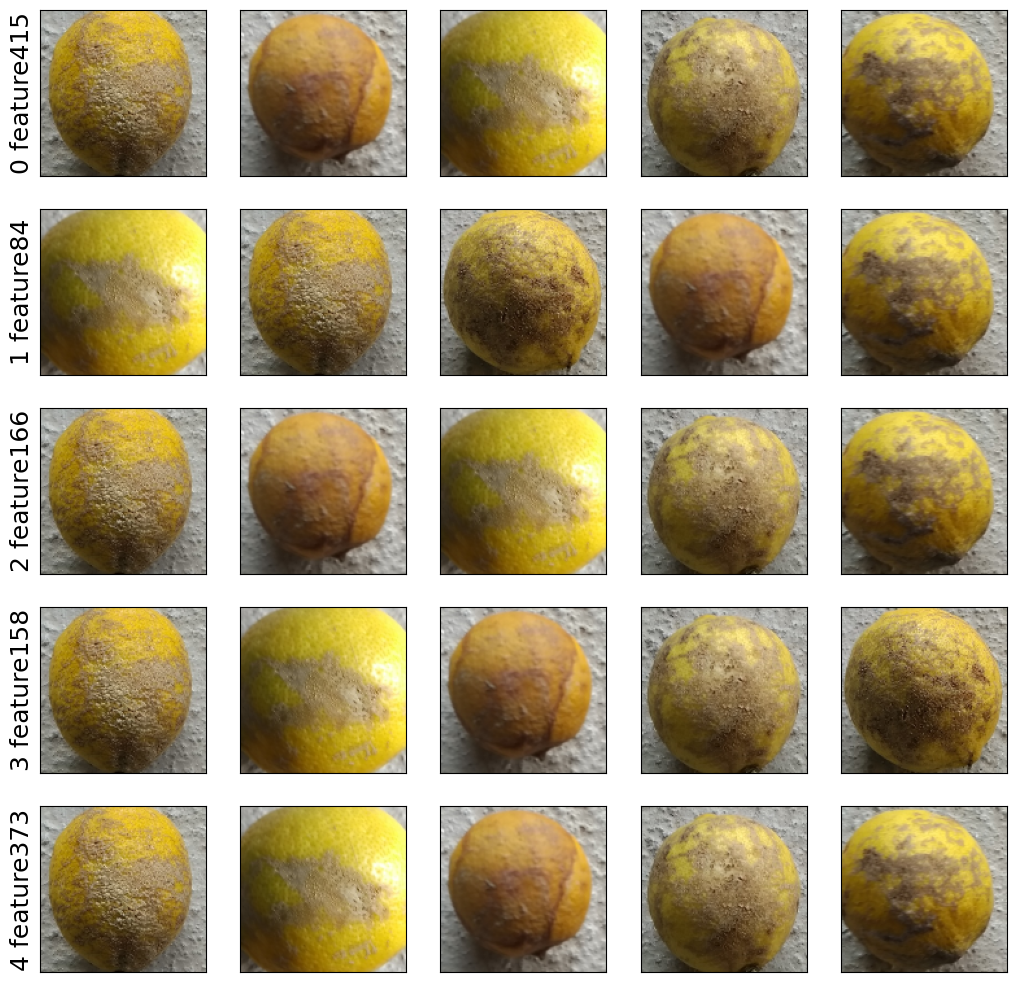}
        \caption{For the 5 most predictive features (from top to bottom) we display the images which yield the highest activations (from left to right).}
    \end{subfigure}
    \hspace{1cm}
    \begin{subfigure}{0.45\textwidth}
        \includegraphics[width=\textwidth]{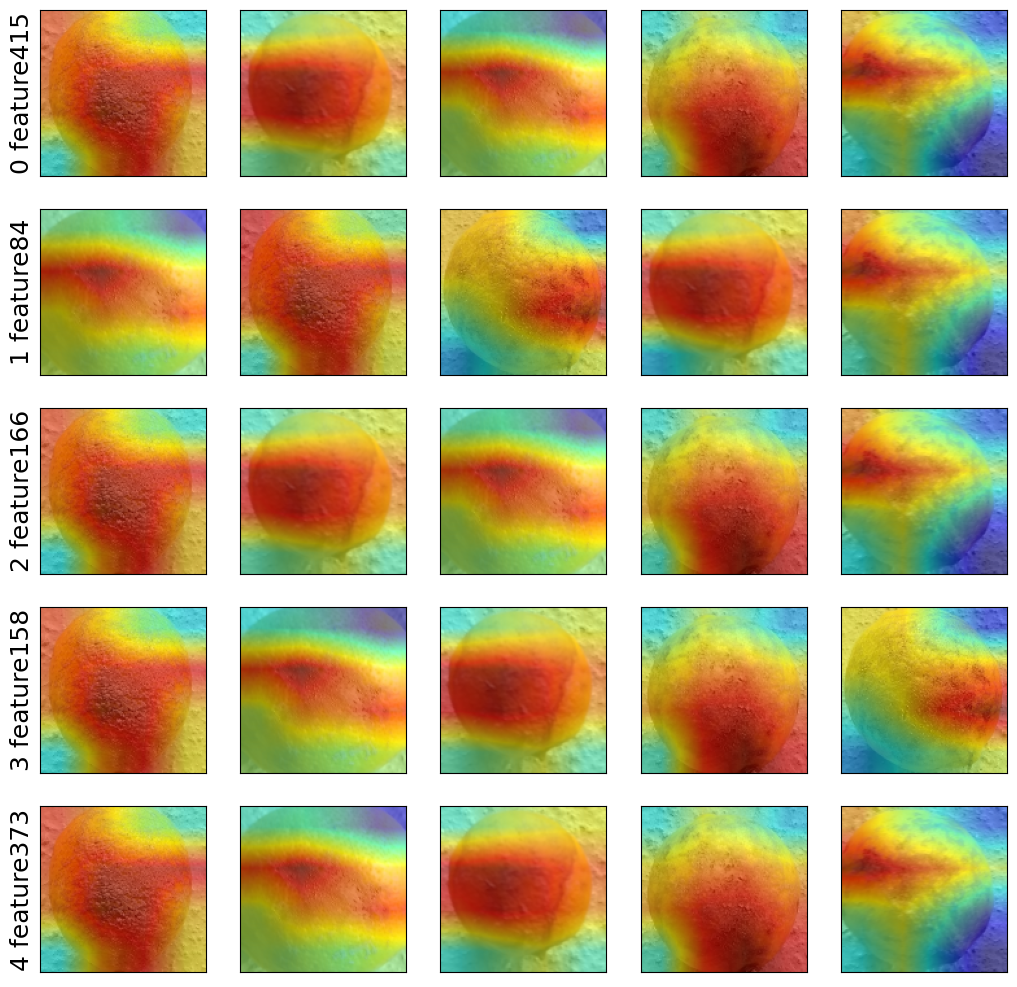}
        \caption{For the 5 most predictive features (from top to bottom) we display the heatmaps on images which yield the highest activations (from left to right).}
    \end{subfigure}
    \caption{Original images along with heatmaps for class `bad quality' of the \emph{Lemon Quality} dataset. A standard CNN classifier identifies the brown patches as a meaningful feature for classification.}
    \label{fig:lemon heatmap class 1}
\end{figure}

\begin{figure}[h!]
\centering
\begin{subfigure}{0.45\textwidth}
\includegraphics[width=\textwidth]{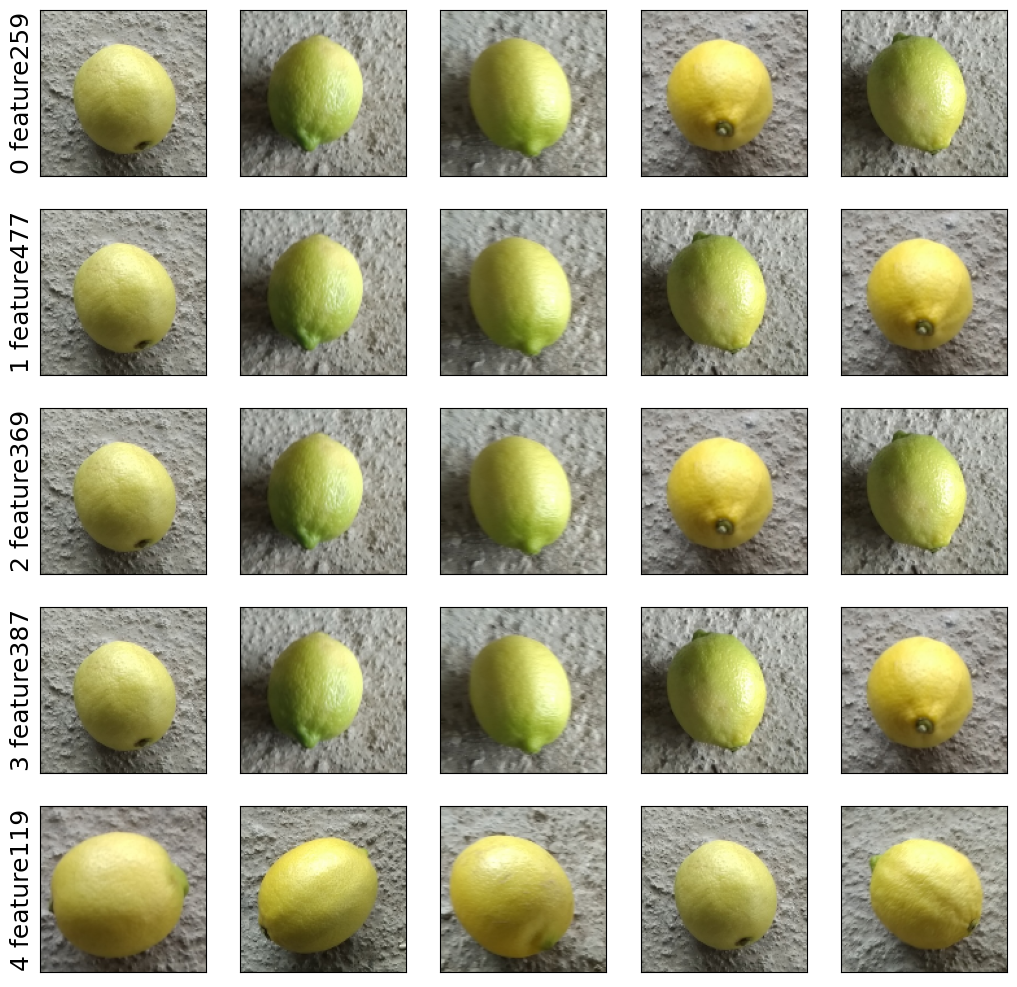}
\caption{For the 5 most predictive features (from top to bottom) we display the images which yield the highest activations (from left to right).}
\end{subfigure}
\hspace{0.5cm}
\begin{subfigure}{0.45\textwidth}
\includegraphics[width=\textwidth]{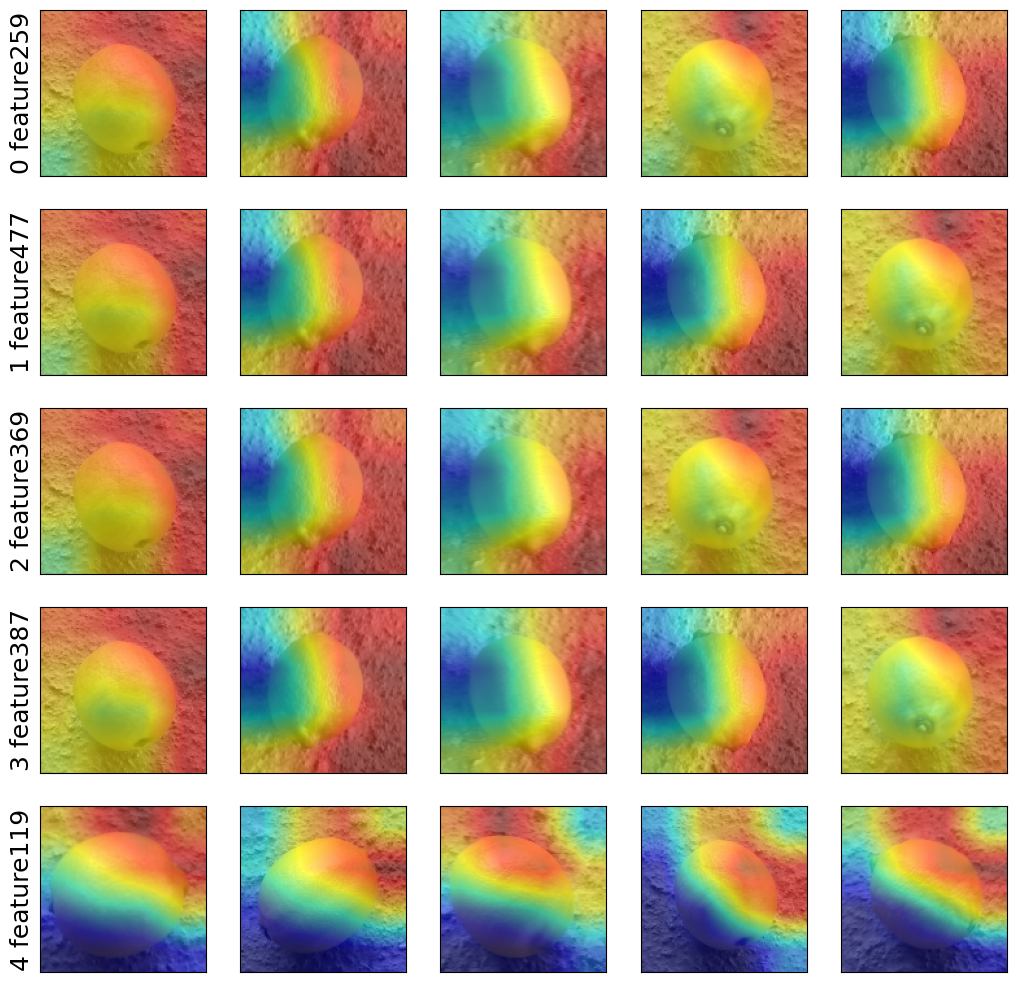}
\caption{For the 5 most predictive features (from top to bottom) we display the heatmaps on images which yield the highest activations (from left to right).}
\end{subfigure}
\caption{Original images along with heatmaps for class `good quality' of the \emph{Lemon Quality} dataset. A standard CNN classifier leverages the background as a spatial shortcut for classification.}
\label{fig:lemon heatmap class 0}
\end{figure}

The heatmaps for the \emph{Lemon Quality} dataset (see \Cref{fig:lemon heatmap class 1}) indicate that the maximally activated features for class `bad quality' are brown patches on the fruit, which is a meaningful feature. However, the heatmaps for class `good quality' (see \Cref{fig:lemon heatmap class 0}) largely focus on the background. Therefore, a heatmap-based approach is able to hint at the spatial shortcut in the \emph{Lemon Quality} dataset. This shortcut is also successfully identified by our method as described in \Cref{ss:results_eval}.

\begin{figure}[h!]
    \centering
    \begin{subfigure}{0.45\textwidth}
        \includegraphics[width=\textwidth]{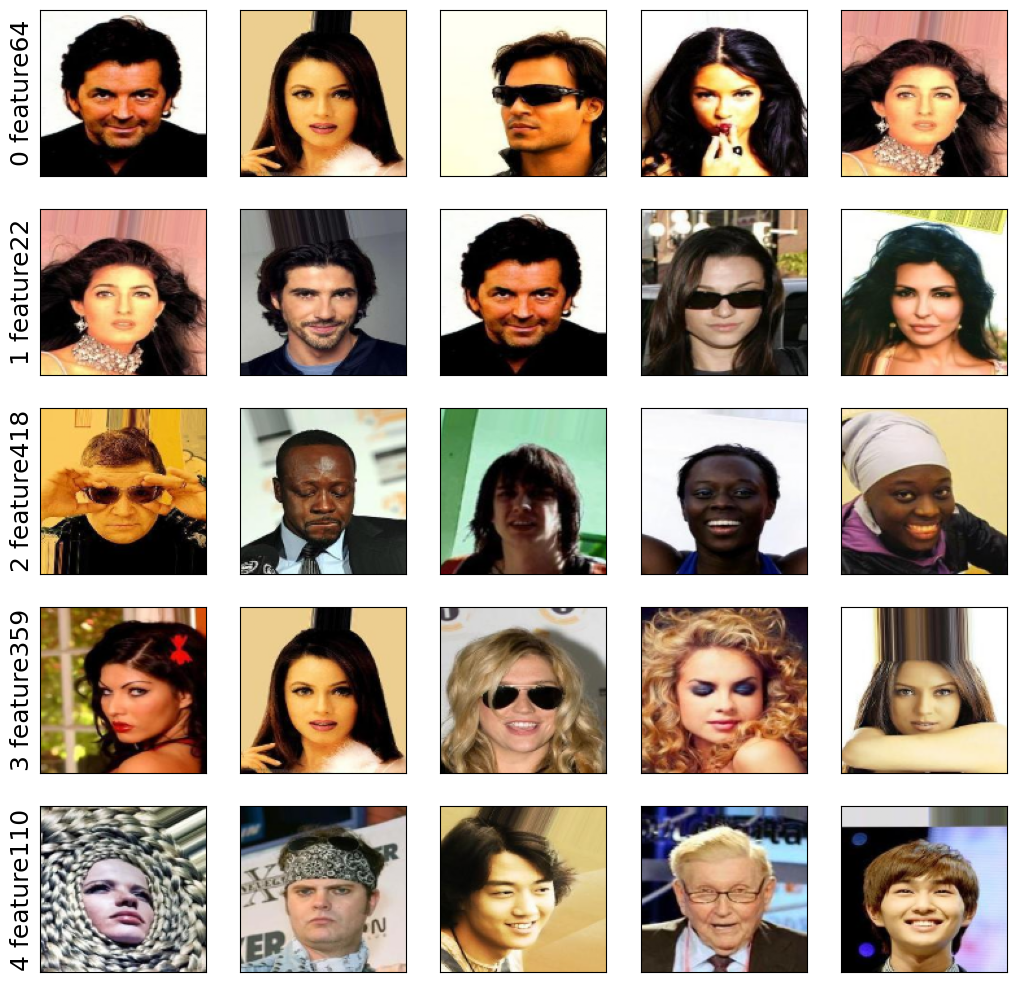}
        \caption{For the 5 most predictive features (from top to bottom) we display the images which yield the highest activations (from left to right).}
    \end{subfigure}
    \hspace{1cm}
    \begin{subfigure}{0.45\textwidth}
        \includegraphics[width=\textwidth]{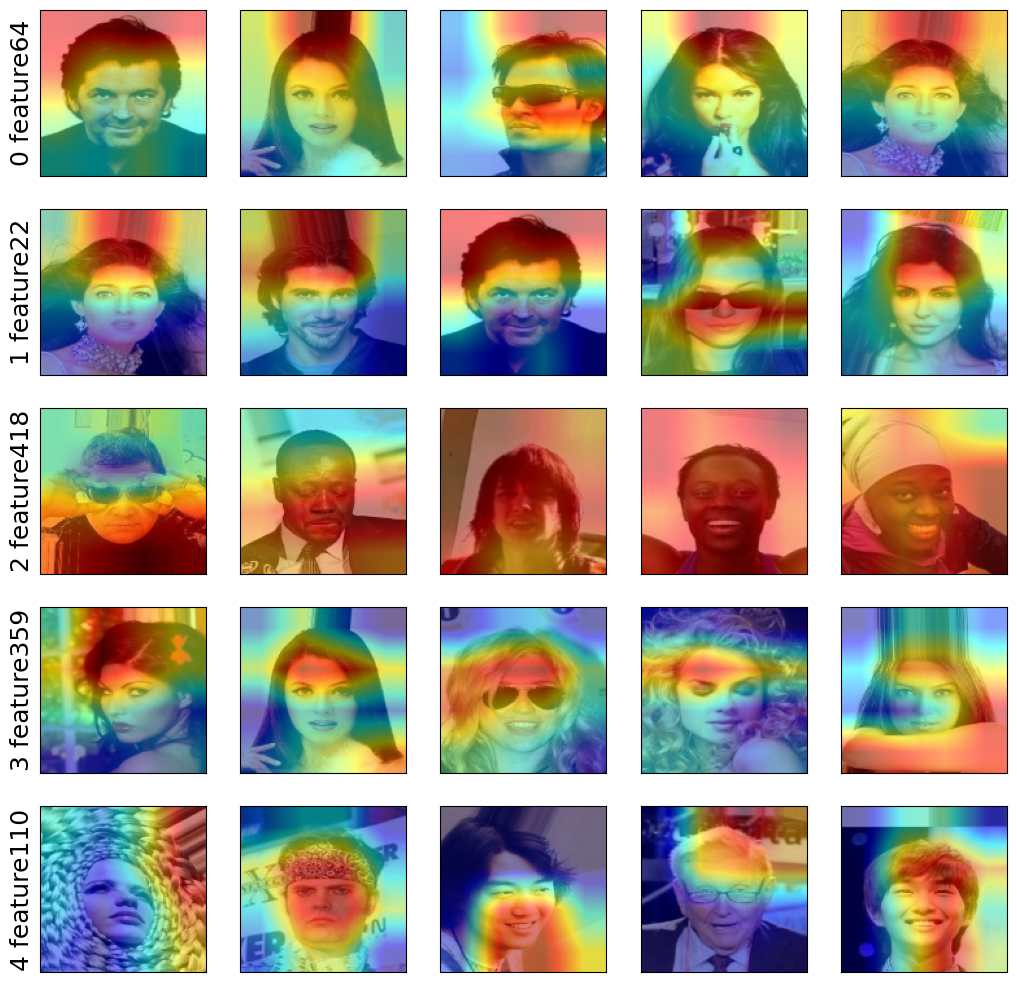}
        \caption{For the 5 most predictive features (from top to bottom) we display the heatmaps on images which yield the highest activations (from left to right).}
    \end{subfigure}
    \caption{Original images along with heatmaps for class `dark hair' of the \emph{CelebA} dataset. While a few heatmaps focus on the valid attribute hair, none of them reveal the gender shortcut.}
    \label{fig:celeba heatmap class 0}
\end{figure}

\begin{figure}[h!]
    \centering
    \begin{subfigure}{0.45\textwidth}
        \includegraphics[width=\textwidth]{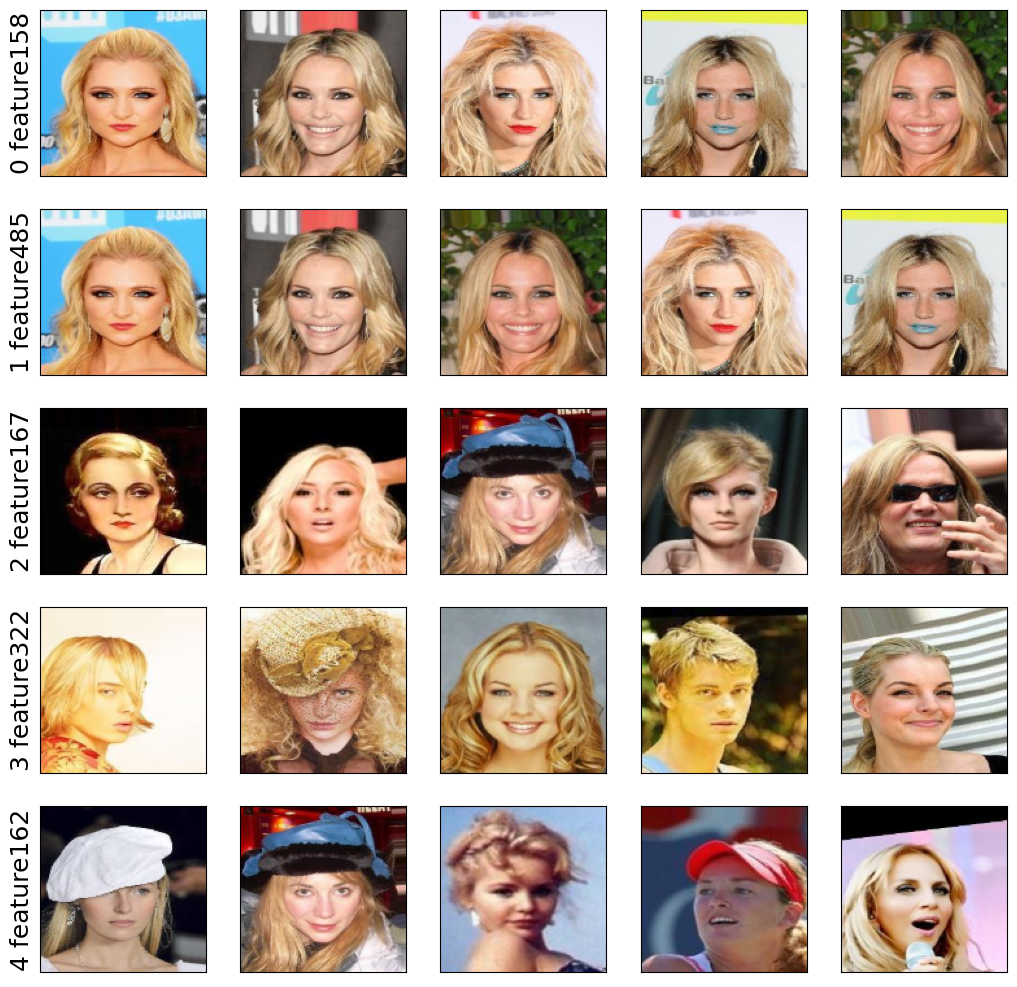}
        \caption{For the 5 most predictive features (from top to bottom) we display the images which yield the highest activations (from left to right).}
    \end{subfigure}
    \hspace{1cm}
    \begin{subfigure}{0.45\textwidth}
        \includegraphics[width=\textwidth]{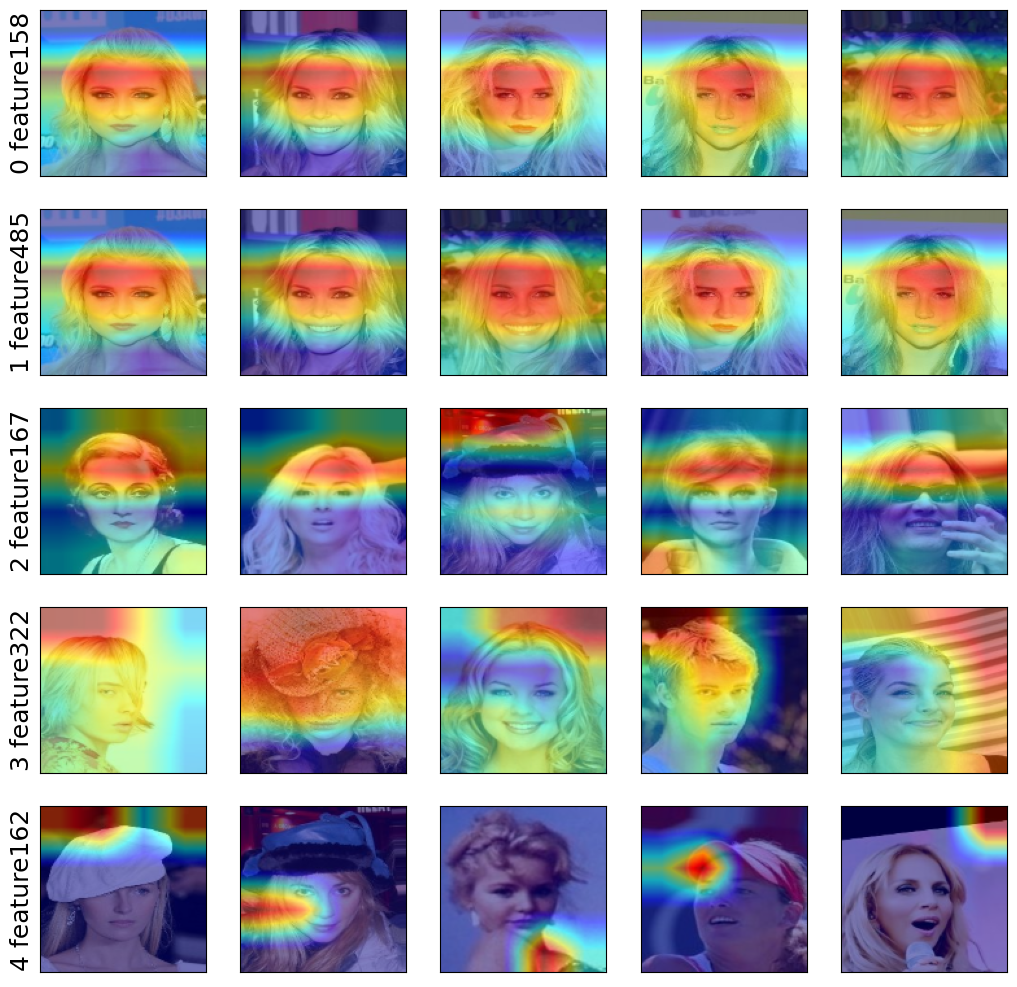}
        \caption{For the 5 most predictive features (from top to bottom) we display the heatmaps on images which yield the highest activations (from left to right).}
    \end{subfigure}
    \caption{Original images along with heatmaps for class `blond hair' of the \emph{CelebA} dataset. The heatmaps highlight meaningful facial features like forehead and eyes but fail to reveal the gender shortcut.}
    \label{fig:celeba heatmap class 1}
\end{figure}

Additionally, we compute the heatmaps on the \emph{CelebA} dataset. The results are illustrated in \Cref{fig:celeba heatmap class 0} and \Cref{fig:celeba heatmap class 1}. 
We can see that the model focuses on the hair in order to classify the `hair color', as well as other facial features such as the forehead and eyes.
However, this does not enable to identify the gender shortcut.
In contrast, our approach could successfully reveal this spurious correlation (see \Cref{fig:celeba_gender_hair_results}).

